\documentclass{article}
\usepackage{spconf,amsmath,graphicx}
\usepackage{amssymb}
\usepackage{url}
\usepackage{booktabs}
\usepackage{balance}
\usepackage{enumitem} 
\usepackage[compress]{cite}

\newcommand{\vect}[1]{\mathbf{#1}}

\title{WIDELY LINEAR KERNELS FOR COMPLEX-VALUED \\ KERNEL ACTIVATION FUNCTIONS}
\name{Simone Scardapane$^{\star}$ \qquad Steven Van Vaerenbergh$^{\dagger}$ \qquad Danilo Comminiello$^{\star}$ \qquad Aurelio Uncini$^{\star}$
\thanks{S. Van Vaerenbergh is supported by the Ministerio de Econom{\'\i}a, Industria y Competitividad (MINECO) of Spain under grant TEC2016-81900-REDT (KERMES). }
}

\address{$^{\star}$ DIET Department, Sapienza University of Rome, Italy \\ $^{\dagger}$Department of Communications Engineering, University of Cantabria, Spain}
\begin{document}
%
\maketitle
\begin{abstract}
Complex-valued neural networks (CVNNs) have been shown to be powerful nonlinear approximators when the input data can be properly modeled in the complex domain. One of the major challenges in scaling up CVNNs in practice is the design of complex activation functions. Recently, we proposed a novel framework for learning these activation functions neuron-wise in a data-dependent fashion, based on a cheap one-dimensional kernel expansion and the idea of kernel activation functions (KAFs). In this paper we argue that, despite its flexibility, this framework is still limited in the class of functions that can be modeled in the complex domain. We leverage the idea of widely linear complex kernels to extend the formulation, allowing for a richer expressiveness without an increase in the number of adaptable parameters. We test the resulting model on a set of complex-valued image classification benchmarks. Experimental results show that the resulting CVNNs can achieve higher accuracy while at the same time converging faster. 
\end{abstract}
\section{Introduction}
\label{sec:intro}

Inference in the complex domain is a fundamental task in both signal processing \cite{schreier2010statistical} and machine learning \cite{hirose2003complex}. Among the approaches proposed over the years, complex-valued neural networks (CVNNs) are gaining a large interest \cite{guberman2016complex,trabelsi2017deep,scardapane2019complex,shafran2018complex}, as they promise to replicate the recent breakthroughs in (real-valued) deep learning to complex-valued problems, such as forecasting and control of complex signals. Working in the complex domain, however, poses a range of unique problems arising from the properties of complex algebra. Foremost among them is the design of complex activation functions \cite{scardapane2019complex}: even extending the rectified linear unit (ReLU) has been shown to be highly non-trivial, with multiple proposals being made over the last two years \cite{guberman2016complex,arjovsky2016unitary}. Several works end up using naive \textit{split} formulations, wherein the real and imaginary parts of the activation are processed independently, with a loss in terms of expressiveness \cite{leung1991complex}.

In \cite{scardapane2019complex} we proposed a different approach, where the activation functions are \textit{learned} in the complex domain via a simple mono-dimensional parameterization. The idea, based on the concept of kernel activation functions (KAFs) originally developed in \cite{scardapane2017kafnets} for the real domain, is to model each function as an independent one-dimensional kernel model, whose mixing weights are adapted through back-propagation, while the dictionary of the kernel matrix is fixed in advance by sampling the complex plane. Despite the empirical performance shown in \cite{scardapane2019complex} on multiple benchmarks problems, in this paper we argue that the expressiveness of each KAF, as defined in \cite{scardapane2019complex}, is still limited when working in the complex domain. In particular, very recently it was shown that the standard formulation of complex-valued kernel methods (which is also adopted in the KAF) is insufficient to model a large set of signals, because more than a single kernel is needed to model the statistics of a complex signal \cite{boloix2017widely,boloix2018complex}. This leads to the concept of \textit{pseudo-kernels} and to \textit{widely linear kernel} methods.

\textit{Contribution of the paper}: in this paper we combine the ideas of \cite{scardapane2019complex} and \cite{boloix2017widely} and we propose a widely linear KAF (WL-KAF) model, a non-parametric activation function defined directly in the complex domain with no constraints on its expressiveness (as opposed to \cite{scardapane2019complex}). We experiment with different choices for the kernel and pseudo-kernel, showing definite improvements on a series of image classification benchmarks in the complex domain, with higher accuracy and faster convergence during optimization.

\textit{Organization of the paper}: in Sections \ref{sec:cvnn} and \ref{sec:cv_afs} we recall the formulation of CVNNs and complex-valued activation functions. Section \ref{sec:proposed_wl_kafs} describes the proposed WL-KAF. Then, we empirically validate its performance in Section \ref{sec:experiments}, before concluding in Section \ref{sec:conclusion} with some remarks on future lines of research.

\section{Complex-valued neural networks}
\label{sec:cvnn}

A CVNN is defined analogously to its real-valued counterpart as the composition of $L$ layers \cite{kim2003approximation}:

\begin{equation}
f(\vect{x}) = \left( f^{L} \circ f^{L-1} \circ \ldots \circ f^1 \right) (\vect{x}) \,,
\label{eq:cvnn}
\end{equation}

\noindent where $\vect{x} \in \mathbb{C}^F$ is the input to the network. Each layer is composed of an adaptable linear projection followed by an element-wise nonlinearity $g$:

\begin{equation}
f^{i}(\vect{h}) = g \left( \vect{W}_i \vect{h} + \vect{b}_{i} \right) \,.
\label{eq:cvnn_layer}
\end{equation}

\noindent where $\vect{W}_i$ and $\vect{b}_i$ are a matrix and a vector that contain (complex-valued) adaptable parameters. While we focus on feedforward networks, we note that by replacing \eqref{eq:cvnn_layer} with more elaborate formulations one can obtain complex equivalents of other types of NNs, e.g., convolutional or recurrent networks \cite{guberman2016complex,trabelsi2017deep,shafran2018complex}. Given $N$ training pairs $\left\{ \vect{x}_n, \vect{y}_n \right\}_{n=1}^N$ we train the network by minimizing a regularized loss:

\begin{equation}
J(\vect{w}) = \sum_{n=1}^N l(\vect{y}_n, f(\vect{x}_n)) + C \cdot \lVert \vect{w} \rVert^2  \,,
\label{eq:global_cost_function}
\end{equation}
where all adaptable parameters are collected in $\vect{w} \in \mathbb{C}^Q$, $l(\cdot, \cdot)$ is a loss function, and $C$ a real-valued scalar (chosen by the user) weighting the regularization term. An example of complex loss is the squared one:

\begin{equation}
l(\vect{y}, \hat{\vect{y}}) = \left(\vect{y} - \hat{\vect{y}}\right)^H\left(\vect{y} - \hat{\vect{y}}\right) \,,
\label{eq:squared_loss}
\end{equation}
where $(\cdot)^H$ is the Hermitian transpose of the vector. Since \eqref{eq:global_cost_function} is non-analytic, CR-calculus \cite{kreutz2009complex,schreier2010statistical} can be used to define proper complex derivatives for use in any optimization algorithm.

\section{Complex-valued activation functions}
\label{sec:cv_afs}

As we stated in the introduction, the design of $g(\cdot)$ in the complex domain is more challenging when compared to the real-valued one, mostly due to Liouville's theorem \cite{hirose2003complex}.\footnote{We only consider the choice of $g(\cdot)$ for the hidden layers, while the choice of the activation function in the outer layer depends on the task (see also Section \ref{sec:experiments}).} It is common for example to work in a split fashion \cite{nitta1997extension}:

\begin{equation}
g(z) = g_R(\Re\left\{z\right\}) + i g_R(\Im\left\{z\right\}) \,,
\label{eq:split_activation_function}
\end{equation}
where $z$ is a single (scalar) activation, $\Re\left\{z\right\}, \Im\left\{z\right\}$ are the real and imaginary components of $z$, and $g_R$ a generic real-valued activation function. Alternative approaches involve phase-amplitude functions acting on the magnitude of the activations, e.g. \cite{georgiou1992complex}:

\begin{equation}
g(z) = \tanh\left\{\lvert z \rvert\right\}\exp\left\{ i\phi(z) \right\} \,,
\end{equation}
where $\phi(z)$ is the phase of $z$. As mentioned in Section \ref{sec:intro}, other authors have also proposed the use of fully complex trigonometric functions, or different variants of the ReLU (commonly used in the real-valued case) \cite{trabelsi2017deep}. We refer to \cite{scardapane2019complex} for a more general overview on the topic. Generally speaking, none of these approaches clearly outperform the others in practice, making it an open research field.

\subsection{Kernel activation functions}

\begin{figure}
\centering
\includegraphics[width=0.6\columnwidth]{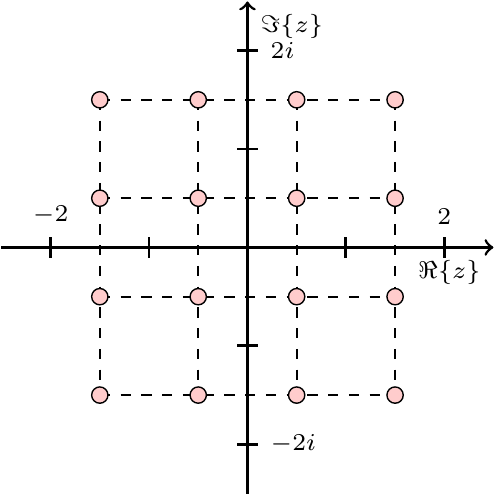}
\caption{Example of dictionary sampling in the complex plane, with $D=16$ elements sampled in $[-2, +2]$ on both axes.}
\label{fig:dictionary_sampling}
\end{figure}

In \cite{scardapane2019complex} we proposed to alleviate the problem of designing complex activation functions by \textit{learning} their shape directly in the complex domain. To this end, we model each activation function (separately for every neuron) with a small number of complex-valued adaptable parameters, representing the linear coefficients in a kernel-based expansion. To introduce the model, we start by sampling the complex space uniformly around $0$, with a resolution chosen by the user, as shown pictorially in Fig. \ref{fig:dictionary_sampling}. The resulting $D$ elements $\vect{d} = \left[d_1, \ldots, d_D\right]^T$ will form our \textit{dictionary}. Given this fixed dictionary, a kernel activation function (KAF) in the complex domain is defined as:\footnote{\cite{scardapane2019complex} also considers a split version of the standard KAF. We focus here on the fully complex extension.}

\begin{equation}
g(z) = \sum_{n=1}^D \alpha_{n} \kappa\left(z, d_n\right) = \vect{k}^T\boldsymbol{\alpha} \,,
\label{eq:complex_kaf}
\end{equation}
where $\kappa$ is a valid kernel function over complex inputs, $\vect{k}$ is a column vector containing the $D$ kernel values computed between $z$ and the dictionary $\vect{d}$, and the parameters $\left\{\alpha_n\right\}_{n=1}^D$ are adapted independently for every neuron, together with the linear weights in \eqref{eq:cvnn_layer}, via standard back-propagation. Fixing the dictionary in advance allows for an extremely efficient (vectorized) implementation of \eqref{eq:complex_kaf} \cite{scardapane2019complex}.

The choice of $\kappa$ can leverage over a large body of literature on complex reproducing kernel Hilbert spaces \cite{steinwart2006explicit,bouboulis2011extension}. In particular, in \cite{scardapane2019complex} we performed experiments with a complex-valued extension of the classical Gaussian kernel:

\begin{equation}
\kappa(z, d) = \exp\left\{-\gamma\left(z - d^* \right)^2\right\} \,,
\label{eq:complex_gaussian_kernel}
\end{equation}
where $\gamma$ is a hyper-parameter, and the independent kernel proposed in \cite{bouboulis2011extension}:

\begin{align}
\kappa\left(z,d\right) & = \kappa_{\mathbb{R}}\left(\Re\left\{z\right\}, \Re\left\{d\right\}\right) + \kappa_{\mathbb{R}}\left(\Im\left\{z\right\}, \Im\left\{d\right\}\right) \nonumber \\
& + i\left( \kappa_{\mathbb{R}}\left(\Re\left\{z\right\}, \Im\left\{d\right\}\right) - \kappa_{\mathbb{R}}\left(\Im\left\{z\right\}, \Re\left\{d\right\}\right)\right) \,.
\label{eq:independent_kernel}
\end{align}
where $\kappa_{\mathbb{R}}$ is a generic real-valued kernel (chosen as the standard Gaussian in \cite{scardapane2019complex}). In the experiments for this paper we will consider a more recent proposal from \cite{boloix2017widely}, a real-valued Gaussian kernel with complex inputs given by:
\begin{equation}
\kappa(z, d) = \exp\left\{-\gamma\left(z - d\right)^*\left(z - d\right)\right\} \,.
\label{eq:complex_gaussian_kernel_2}
\end{equation}

\section{Proposed widely linear KAF}
\label{sec:proposed_wl_kafs}
The key motivation for this paper is that the model in \eqref{eq:complex_kaf} is limited in the kind of complex-valued function it can approximate, an observation first made in \cite{boloix2017widely}. To see this, note that one can express the complex function $g(z)$ in terms of a kernel method with two outputs, namely, the real and imaginary parts $g_r(z)$, $g_i(z)$. According to the theory of vector-valued kernel methods \cite{alvarez2012kernels}, the corresponding kernel is now \textit{matrix-valued} and the output can be written as: 
\begin{equation}
g(z) = 
\begin{bmatrix}
g_r(z) \\ g_i(z) 
\end{bmatrix} =
\begin{bmatrix}
\vect{k}^T_{rr} & \vect{k}^T_{ri} \\
\vect{k}^T_{ir} & \vect{k}^T_{ii}
\end{bmatrix}
\begin{bmatrix}
\boldsymbol{\alpha}_r \\ \boldsymbol{\alpha}_i
\end{bmatrix} \,,
\label{eq:vector_model}
\end{equation}
where we now have \textit{four} column vectors $\left\{\vect{k}_{rr}, \vect{k}_{ri}, \vect{k}_{ir}, \vect{k}_{ii}\right\}$ corresponding to the four outputs of the kernel , and two sets of linear weights $\boldsymbol{\alpha}_r$ and $\boldsymbol{\alpha}_i$. Substituting \eqref{eq:complex_kaf} into \eqref{eq:vector_model} shows that \eqref{eq:complex_kaf} forces the constraints $\vect{k}_{rr} = \vect{k}_{ii}$ and $\vect{k}_{ri} = - \vect{k}_{ir}$, limiting the expressiveness of the overall model. A solution to this is the adoption of widely linear kernel methods \cite{boloix2017widely}.

Following this, we propose an extension of the complex-valued KAF adopting widely linear kernels, that we term widely linear KAF (WL-KAF):
\begin{equation}
g(z) = \vect{k}^T\boldsymbol{\alpha} + \widetilde{\vect{k}}^T\boldsymbol{\alpha}^* \,,
\label{eq:proposed_wl_kaf}
\end{equation}
where $\widetilde{\vect{k}} = \left[ \widetilde{\kappa}(z, d_1), \ldots, \widetilde{\kappa}(z, d_D) \right]$, and $\widetilde{\kappa}$ is called the `pseudo-kernel'. $\boldsymbol{\alpha}^*$ is the complex conjugate of $\boldsymbol{\alpha}$. The model in \eqref{eq:proposed_wl_kaf} does not impose the previously discussed limitations, and it can be shown that:
\begin{eqnarray}
\vect{k} & = 0.5 \left[ \vect{k}_{rr} + \vect{k}_{ii} + i\left( \vect{k}_{ir} - \vect{k}_{ri} \right)\right] \,, \\
\widetilde{\vect{k}} & = 0.5 \left[ \vect{k}_{rr} - \vect{k}_{ii} + i\left( \vect{k}_{ir} + \vect{k}_{ri} \right)\right] \,.
\end{eqnarray}
Depending on the choice of the kernel and pseudo-kernel, the resulting model has a larger amount of expressiveness compared to the standard one. In the context of KAFs and CVNNs, the model has two additional properties to its favor. Firstly, as we will see shortly, since the dictionary is fixed the kernel and pseudo-kernel can generally share a large amount of computation, making the modification extremely cheap in terms of speed. Secondly, the use of widely linear models does not increase the number of adaptable parameters, since in our case we are only adapting the mixing coefficients $\boldsymbol{\alpha}$. Following \cite{boloix2017widely}, in the experiments we consider two different choices for the kernel and pseudo-kernel.

\textbf{Case 1}: if we assume that the real and imaginary parts of $g(z)$ are independent, the off-diagonal blocks in \eqref{eq:vector_model} cancel and we are left with:
\begin{eqnarray}
\vect{k} & = 0.5 \left[ \vect{k}_{rr} + \vect{k}_{ii} \right] \,, \\
\widetilde{\vect{k}} & = 0.5 \left[ \vect{k}_{rr} - \vect{k}_{ii} \right] \,.
\end{eqnarray}
In this case we use \eqref{eq:complex_gaussian_kernel_2} with two separate parameters $\gamma$ for $\vect{k}_{rr}$ and $\vect{k}_{ii}$. More specifically, both bandwidths in our experiments are initialized following the rule of thumb taken from \cite{scardapane2017kafnets}, but are subsequently adapted via back-propagation independently for every neuron.

\textbf{Case 2}: in the case where the real and imaginary parts are not assumed independent, we can exploit the theory of separable kernels and mixed effect regularizers introduced for vector-valued kernels \cite{alvarez2012kernels}. In our case we obtain, for an hyper-parameter $Q$ chosen by the user \cite{boloix2017widely}:
\begin{align}
\kappa(z, d) & = \sum_{q=1}^Q \kappa^q(z, d) \,,\\
\widetilde{\kappa}(z, d) & = 2i \sum_{q=1}^Q \omega^q \widetilde{\kappa}^q(z, d) \,,
\label{eq:case_2}
\end{align}
with all the kernels $\kappa^q$ and $\widetilde{\kappa}^q$ being real-valued in output, and $0 < \omega^q < 1$. As before, one can exploit different Gaussian kernels as in \eqref{eq:complex_gaussian_kernel_2}, letting the different bandwidths adapt via back-propagation.

\section{Experimental evaluation}
\label{sec:experiments}

\begin{table*}[!ht]
\caption{Test accuracy (mean and standard deviation) for the complex-valued image classification benchmarks (see main discussion for the preprocessing phase). First two rows are taken from \cite{scardapane2019complex}. The best results for each dataset are highlighted in bold.}
{\centering\hfill{}
	\setlength{\tabcolsep}{4pt}
	\renewcommand{\arraystretch}{1.5}
	\begin{normalsize}
	\begin{tabular}{lcccc}   
	\toprule
	\textbf{Model} & \textbf{MNIST} & \textbf{F-MNIST} & \textbf{E-MNIST} & \textbf{Latin OCR} \\ 
	\midrule
	Real-valued NN & $92.39 \pm 0.10$ & $71.08 \pm 0.45$ & $92.78 \pm 1.25$ & $39.01 \pm 3.42$ \\
	KAF & $97.18 \pm 0.27$ & $81.94 \pm 0.91$ & $98.11 \pm 2.04$ & $71.79 \pm 2.40$ \\
	\midrule
	\textbf{Proposed WL-KAF (Case 1)} & $\vect{97.50 \pm 0.41}$ & $77.29 \pm 2.43$ & $98.46 \pm 0.12$ & $\vect{74.57 \pm 0.80}$ \\
	\textbf{Proposed WL-KAF (Case 2)} & $96.22 \pm 0.74$ & $\vect{82.89 \pm 1.09}$ & $\vect{99.03 \pm 1.01}$ & $72.53 \pm 0.36$ \\
	\bottomrule
	\end{tabular}
	\end{normalsize}
}
\hfill{}
\label{tab:results}
\end{table*}

We evaluate the two proposed WL-KAFs on a series of complex-valued image classification benchmarks extended from \cite{scardapane2019complex}. We consider four problems: 

\begin{itemize}[noitemsep]
\item \textbf{MNIST},\footnote{\url{http://yann.lecun.com/exdb/mnist/}} composed of $60000$ $28 \times 28$ images belonging to ten digit classes. 
\item \textbf{Fashion MNIST} (F-MNIST) \cite{xiao2017fashion}: a variant of MNIST where classes are clothing items, with the same dimensionality and size as MNIST.
\item \textbf{Extended MNIST} (EMNIST) \cite{cohen2017emnist}: we use the `Digits' extension, having $240$ thousand images of handwritten digits.
\item \textbf{Latin OCR} \cite{firmani2017codice}: an OCR problem of handwritten Latin characters extracted from manuscripts of the Vatican secret archives. There are $12000$ 
images and $23$ classes.
\end{itemize}
To convert these to complex-valued problems, we adopt the procedure from \cite{bouboulis2015complex} and preprocess each image with a fast Fourier transform (FFT), then rank the coefficients of the FFT in terms of significance (by considering their mean absolute value), keeping only the $100$ most significant coefficients as input to the models.

The results in \cite{scardapane2019complex} are taken as a baseline, to which we add two CVNNs of the same dimensionality as \cite{scardapane2019complex} (three hidden layers of $100$ neurons each) exploiting the proposed WL-KAF. We use a dictionary by sampling $8$ points on each axis equispaced in $\left[-2, +2\right]$. For the case 2 in \eqref{eq:case_2}, as in \cite{boloix2017widely}, we use $Q=1$, $\omega^1 = 0.3$, and the Gaussian kernel in \eqref{eq:complex_gaussian_kernel_2} for the two kernels. As stated before, in all cases the kernel bandwidth $\gamma$ in \eqref{eq:complex_gaussian_kernel_2} is initialized with the rule of thumb in \cite{scardapane2017kafnets} and then adapted independently for every kernel via backpropagation.  The KAFs are applied only to intermediate layers, while the output $\vect{h}$ of the last linear projection is fed to a softmax-like function to compute the class probabilities:
\begin{equation}
\text{softmax}_n(\vect{h}) = \frac{\exp\left\{\Re\left\{h_n\right\}^2+\Im\left\{h_n\right\}^2\right\}}{\sum_{t=1}^C \exp\left\{\Re\left\{h_t\right\}^2+\Im\left\{h_t\right\}^2\right\}} \,,
\end{equation} 
We minimize a regularized cross-entropy over the training data, where the amount of regularization is found through grid search as in \cite{scardapane2019complex}. We use a version of the Adagrad algorithm on random mini-batches of $40$ images to perform optimization. We further employ an early stopping procedure, stopping the optimization whenever the accuracy computed over the validation split of the dataset is not improving for $1000$ iterations of optimization.

\begin{figure}
\centering
\includegraphics[width=0.95\columnwidth]{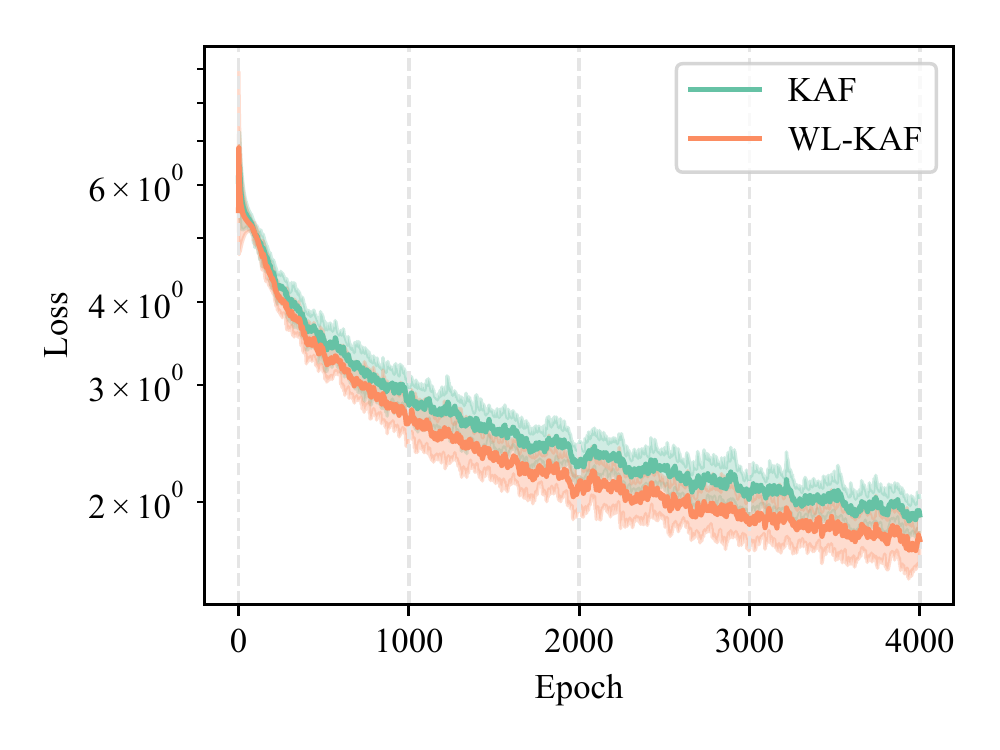}
\caption{Convergence of KAF and WL-KAF (case 1) on the Latin OCR dataset. Standard deviation is shown with a lighter color, while the plot is zoomed on the first 4000 iterations.}
\label{fig:convergence}
\end{figure}

The results of the experiments are provided in Table \ref{tab:results}. ``Real-valued NN'' is a NN having the same dimensionality as the others, but treating real and imaginary parts of the input vector as separate inputs. ``KAF'' is the KAF in \eqref{eq:complex_kaf} using the independent kernel in \eqref{eq:independent_kernel}. As can be seen, CVNNs with the proposed WL-KAFs can achieve in all cases a superior performance, without introducing additional parameters compared to the standard complex-valued KAF. This increase in performance translates to faster convergence, an example of which (on the Latin OCR dataset) is shown in Fig. \ref{fig:convergence}.

\section{Conclusion}
\label{sec:conclusion}

In this paper we proposed a new model for learning activation functions for complex-valued neural networks. The model extends the idea of kernel activation functions (KAFs), by incorporating recent ideas from the field of widely linear kernel approximation. Compared to the standard KAF, the widely linear KAF does not require additional trainable parameters while possessing increased flexibility. On a set of complex-valued image classification benchmarks, it achieves better accuracy in all problems while at the same time being faster in terms of optimization. Future work will consider a formal analysis of the generalization properties of the proposed KAFs, and their evaluation in more elaborate complex benchmarks. For the latter, we plan a more comprehensive evaluation of kernels over complex spaces, along with the definition of proper strategies for finding complex hyperparameters (e.g., complex-valued learning rates in the optimization procedure \cite{zhang2016complex}).



\balance
\bibliographystyle{IEEEbib}
\bibliography{refs}

\begin{thebibliography}{10}

\bibitem{schreier2010statistical}
P.~J. Schreier and L.~L. Scharf,
\newblock {\em Statistical signal processing of complex-valued data: the theory
  of improper and noncircular signals},
\newblock Cambridge University Press, 2010.

\bibitem{hirose2003complex}
A.~Hirose,
\newblock {\em Complex-valued neural networks: theories and applications},
  vol.~5,
\newblock World Scientific, 2003.

\bibitem{guberman2016complex}
N.~Guberman,
\newblock ``On complex valued convolutional neural networks,''
\newblock {\em arXiv preprint arXiv:1602.09046}, 2016.

\bibitem{trabelsi2017deep}
C.~Trabelsi, O.~Bilaniuk, D.~Serdyuk, S.~Subramanian, J.~F. Santos, S.~Mehri,
  N.~Rostamzadeh, Y.~Bengio, and C.~J. Pal,
\newblock ``Deep complex networks,''
\newblock {\em 35th International Conference on Machine Learning (ICML)}, 2018.

\bibitem{scardapane2019complex}
S.~Scardapane, S.~Van~Vaerenbergh, A.~Hussain, and A.~Uncini,
\newblock ``Complex-valued neural networks with non-parametric activation
  functions,''
\newblock {\em IEEE Transactions on Emerging Topics in Computational
  Intelligence}, 2019,
\newblock in press.

\bibitem{shafran2018complex}
Izhak Shafran, Tom Bagby, and RJ~Skerry-Ryan,
\newblock ``Complex evolution recurrent neural networks (cernns),''
\newblock in {\em 2018 IEEE International Conference on Acoustics, Speech and
  Signal Processing (ICASSP)}. IEEE, 2018, pp. 5854--5858.

\bibitem{arjovsky2016unitary}
M.~Arjovsky, A.~Shah, and Y.~Bengio,
\newblock ``Unitary evolution recurrent neural networks,''
\newblock in {\em 33rd International Conference on Machine Learning (ICML)},
  2016, pp. 1120--1128.

\bibitem{leung1991complex}
H.~Leung and S.~Haykin,
\newblock ``The complex backpropagation algorithm,''
\newblock {\em IEEE Transactions on Signal Processing}, vol. 39, no. 9, pp.
  2101--2104, 1991.

\bibitem{scardapane2017kafnets}
S.~Scardapane, S.~Van~Vaerenbergh, S.~Totaro, and A.~Uncini,
\newblock ``Kafnets: kernel-based non-parametric activation functions for
  neural networks,''
\newblock {\em Neural Networks}, vol. 110, pp. 19--32, 2019.

\bibitem{boloix2017widely}
R.~Boloix-Tortosa, J.~J. Murillo-Fuentes, I.~Santos, and F.~P{\'e}rez-Cruz,
\newblock ``Widely linear complex-valued kernel methods for regression,''
\newblock {\em IEEE Transactions on Signal Processing}, vol. 65, no. 19, pp.
  5240--5248, 2017.

\bibitem{boloix2018complex}
R.~Boloix-Tortosa, J.~J. Murillo-Fuentes, F.~J. Pay{\'a}n-Somet, and
  F.~P{\'e}rez-Cruz,
\newblock ``Complex {G}aussian processes for regression,''
\newblock {\em IEEE Transactions on Neural Networks and Learning Systems},
  2018.

\bibitem{kim2003approximation}
T.~Kim and T.~Adal{\i},
\newblock ``Approximation by fully complex multilayer perceptrons,''
\newblock {\em Neural Computation}, vol. 15, no. 7, pp. 1641--1666, 2003.

\bibitem{kreutz2009complex}
K.~Kreutz-Delgado,
\newblock ``The complex gradient operator and the {CR}-calculus,''
\newblock {\em arXiv preprint arXiv:0906.4835}, 2009.

\bibitem{nitta1997extension}
T.~Nitta,
\newblock ``An extension of the back-propagation algorithm to complex
  numbers,''
\newblock {\em Neural Networks}, vol. 10, no. 8, pp. 1391--1415, 1997.

\bibitem{georgiou1992complex}
G.~M. Georgiou and C.~Koutsougeras,
\newblock ``Complex domain backpropagation,''
\newblock {\em IEEE Transactions on Circuits and Systems II: Analog and Digital
  Signal Processing}, vol. 39, no. 5, pp. 330--334, 1992.

\bibitem{steinwart2006explicit}
I.~Steinwart, D.~Hush, and C.~Scovel,
\newblock ``An explicit description of the reproducing kernel {H}ilbert spaces
  of {G}aussian {RBF} kernels,''
\newblock {\em IEEE Transactions on Information Theory}, vol. 52, no. 10, pp.
  4635--4643, 2006.

\bibitem{bouboulis2011extension}
P.~Bouboulis and S.~Theodoridis,
\newblock ``Extension of {Wirtinger}'s calculus to reproducing kernel {Hilbert}
  spaces and the complex kernel {LMS},''
\newblock {\em IEEE Transactions on Signal Processing}, vol. 59, no. 3, pp.
  964--978, 2011.

\bibitem{alvarez2012kernels}
M.~A. Alvarez, L.~Rosasco, and N.~D. Lawrence,
\newblock ``Kernels for vector-valued functions: A review,''
\newblock {\em Foundations and Trends{\textregistered} in Machine Learning},
  vol. 4, no. 3, pp. 195--266, 2012.

\bibitem{xiao2017fashion}
H.~Xiao, K.~Rasul, and R.~Vollgraf,
\newblock ``Fashion-{MNIST}: a novel image dataset for benchmarking machine
  learning algorithms,''
\newblock {\em arXiv preprint arXiv:1708.07747}, 2017.

\bibitem{cohen2017emnist}
G.~Cohen, S.~Afshar, J.~Tapson, and A.~van Schaik,
\newblock ``{EMNIST}: an extension of {MNIST} to handwritten letters,''
\newblock {\em arXiv preprint arXiv:1702.05373}, 2017.

\bibitem{firmani2017codice}
D.~Firmani, P.~Merialdo, E.~Nieddu, and S.~Scardapane,
\newblock ``In {C}odice {R}atio: {OCR} of handwritten latin documents using
  deep convolutional networks,''
\newblock in {\em 11th International Workshop on Artificial Intelligence for
  Cultural Heritage (AI*CH 2017)}. CEUR Workshop Proceedings, 2017, pp. 9--16.

\bibitem{bouboulis2015complex}
P.~Bouboulis, S.~Theodoridis, C.~Mavroforakis, and L.~Evaggelatou-Dalla,
\newblock ``Complex support vector machines for regression and quaternary
  classification,''
\newblock {\em IEEE Transactions on Neural Networks and Learning Systems}, vol.
  26, no. 6, pp. 1260--1274, 2015.

\bibitem{zhang2016complex}
H.~Zhang and D.~P. Mandic,
\newblock ``Is a complex-valued stepsize advantageous in complex-valued
  gradient learning algorithms?,''
\newblock {\em IEEE Transactions on Neural Networks and Learning Systems}, vol.
  27, no. 12, pp. 2730--2735, 2016.

\end{thebibliography}

\end{document}